\documentclass{new_tlp}
\usepackage{mathptmx}
\usepackage{times}
\usepackage{amssymb}
\usepackage{amsfonts}
\usepackage{latexsym}
\usepackage{url}
\usepackage{epsfig}
\usepackage{amsfonts}
\usepackage{amsmath}
\usepackage{bm}
\usepackage{wrapfig}
\newfont{\Bb}{msbm10}
\newtheorem{theorem}{Theorem}
\newtheorem{proposition}{Proposition}

\newtheorem{example}{Example}
\newtheorem{definition}{Definition}
\newcommand*\rfrac[2]{{}^{#1}\!/_{#2}}
\newcommand{\ignore}[1]{}

\newcommand{\Lower}[1]{\smash{\lower 1.5ex \hbox{#1}}}

\title{Partial Evaluation of Logic Programs in Vector Spaces}
      
\author[C. Sakama et al.]
      {CHIAKI SAKAMA\\ Wakayama University, Japan\\ \email{sakama@sys.wakayama-u.ac.jp}
\and 
      HIEN D. NGUYEN\\ University of Information Technology, VNU-HCM, Vietnam\\ \email{hiennd@uit.edu.vn}
\and 
      TAISUKE SATO\\  AI Research Center AIST, Japan\\ \email{satou.taisuke@aist.go.jp}
\and
      KATSUMI INOUE\\ National Institute of Informatics (NII), Japan\\ \email{inoue@nii.ac.jp}}

\submitted{15 April 2018}
\revised{27 May 2018}
\accepted{27 May 2018}

\begin{document}
\label{firstpage}

\maketitle

\begin{abstract}
In this paper, we introduce methods of encoding propositional logic programs in vector spaces. 
Interpretations are represented by vectors and programs are represented by matrices. 
The least model of a definite program is computed by multiplying an interpretation vector and 
a program matrix.  
To optimize computation in vector spaces, we provide a method of partial evaluation of programs using linear algebra.  Partial evaluation is done by unfolding rules in a program, and it is realized 
in a vector space by multiplying program matrices. 
We perform experiments using randomly generated programs and show that partial evaluation 
has potential for realizing efficient computation in huge scale of programs. 
\end{abstract}



\section{Introduction} \label{sec:1}

One of the challenging topics in AI is to reason with huge scale of knowledge bases.  
Linear algebraic computation has potential to make symbolic reasoning scalable to real-life 
datasets, and several studies aim at integrating 
linear algebraic computation and symbolic computation. 
For instance, Grefenstette \citeyear{Gre13} introduces tensor-based predicate calculus that 
realizes logical operations. 
Yang, $et\, al.$ \citeyear{YYH15} introduce a method of mining Horn clauses from 
relational facts represented in a vector space. 
Serafini and Garcez \citeyear{SG16} introduce logic tensor networks that integrate  
logical deductive reasoning and data-driven relational learning. 
Sato \citeyear{Sato17a} formalizes Tarskian semantics of first-order logic in vector spaces, and  
\cite{Sato17b} shows that tensorization realizes efficient computation of Datalog.  
Lin \citeyear{Lin13} introduces linear algebraic computation of SAT for clausal theories. 

To realize linear algebraic computation of logic programming,   
Sakama $et\,al.$ \citeyear{SIS17} introduce encodings of Horn, disjunctive and normal logic programs 
in vector spaces.  They show that  
least models of Horn programs, minimal models of disjunctive programs, and stable models of 
normal programs are computed by algebraic manipulation of third-order tensors. 
The study builds a new theory of logic programming, while implementation 
and evaluation are left open. 

In this paper, we first reformulate the framework of \cite{SIS17} and present an algorithm for 
computing least models of definite programs in vector spaces. 
We next introduce two optimization techniques for computing: the first one is based on 
column reduction of matrices, and the second one is based on {\em partial evaluation}.  
We perform experimental testing and 
compare algorithms for computing fixpoints of definite programs. 
The rest of this paper is organized as follows. 
Section~2 reviews basic notions and Section~3 provides 
linear algebraic characterization of logic programming.  
Section~4 presents partial evaluation of logic programs in vector spaces. 
Section~5 provides experimental results and Section~6 summarizes the paper. 
Due to space limit, we omit proofs of propositions and theorems. 

\section{Preliminaries}  \label{sec:2}

We consider a language ${\cal L}$ that contains a finite set of propositional variables. 
Given a logic program $P$, the set of all propositional variables appearing in $P$ is called 
the {\em Herbrand base\/} of $P$ (written $B_P$). 
A {\em definite program\/} is a finite set of {\em rules\/} of the form: 
\begin{equation} \label{h-rule}
h\leftarrow\; b_1\wedge\cdots\wedge b_m\;\;\;\;  (m\geq 0)
\end{equation}
where $h$ and $b_i$ are propositional variables (atoms) in ${\cal L}$. 
A rule $r$ is called a {\em d-rule\/} if $r$ is the form: 
\begin{equation} \label{d-rule}
h\leftarrow\; b_1\vee\cdots\vee b_m\;\;\;\;  (m\geq 0)
\end{equation}
where $h$ and $b_i$ are propositional variables in ${\cal L}$. 
A {\em d-program\/} is a finite set of rules that are either~(\ref{h-rule}) or~(\ref{d-rule}). 
Note that the rule~(\ref{d-rule}) is a shorthand of $m$ rules: $h\leftarrow b_1$, $\ldots$, $h\leftarrow b_m$, so a d-program is considered a definite program.%
\footnote{The notion of d-programs is useful when we consider a program such that each atom is 
defined by a single rule in Section~\ref{sec:3}.}
For each rule $r$ of the form~(\ref{h-rule}) or~(\ref{d-rule}),  
define $head(r)=h$ and $body(r)=\{b_1,\ldots, b_m\}$.%
\footnote{We assume $b_i\neq b_j$ if $i\neq j$.}
A rule is called a {\em fact\/} if $body(r)=\emptyset$. 

A set $I\subseteq B_P$ is an {\em interpretation\/} of $P$. 
An interpretation $I$ is a {\em model\/} of a d-program $P$ 
if $\{b_1,\ldots,b_m\}\subseteq I$ implies $h\in I$ for 
every rule~(\ref{h-rule}) in $P$, and  $\{b_1,\ldots,b_m\}\cap I\neq\emptyset$ implies $h\in I$ for 
every rule~(\ref{d-rule}) in $P$. 
A model $I$ is the {\em least model\/} of $P$ if $I\subseteq J$ for any model $J$ of $P$. 
A mapping $T_P:\, 2^{B_P} \rightarrow 2^{B_P}$ (called a $T_P$-{\em operator\/}) is defined as: 
\begin{eqnarray*}
&& T_P(I) = \{\, h\,\mid\, h\leftarrow b_1\wedge\cdots\wedge b_m\in P\;\mbox{and}\; 
                \{ b_1,\ldots, b_m \}\subseteq I\,\}\\
&& \qquad\qquad \cup\; \{\, h\,\mid\, h\leftarrow b_1\vee\cdots\vee b_n\in P\;\mbox{and}\; 
                \{ b_1,\ldots, b_n \}\cap I\neq\emptyset\,\}. 
\end{eqnarray*}
The {\em powers\/} of $T_P$ are defined as: 
$T_P^{k+1}(I)=T_P(T_P^k(I))$ $(k\geq 0)$ and $T_P^0(I)=I$. 
Given $I\subseteq B_P$, there is a fixpoint $T_P^{n+1}(I)=T_P^n(I)$ $(n\geq 0)$. 
For a definite program $P$, the fixpoint $T_P^n(\emptyset)$ coincides with the least model of $P$ 
\cite{vEK76}.  


\section{Logic Programming in Linear Algebra}  \label{sec:3}
\subsection{SD programs}  \label{sec:3.1}
 
We first consider a subclass of definite programs, called SD programs. 

\begin{definition}[SD program]\rm 
A definite program $P$ is called {\em singly defined\/} ($SD$ {\em program\/}, for short) if 
$head(r_1)\neq head(r_2)$ for any two rules $r_1$ and $r_2$ ($r_1\neq r_2$) in $P$. 
\end{definition}

Interpretations and programs are represented in a vector space as follows.  

\begin{definition}[interpretation vector~\cite{SIS17}]\rm
Let $P$ be a definite program and $B_P=\{ p_1,\ldots,p_n \}$. 
Then an interpretation $I\subseteq B_P$ is represented by a vector 
$\bm{v}=(a_1,\ldots,a_n)^{\sf T}$ where 
each element $a_i$ $(1\leq i\leq n)$ represents the truth value of the proposition $p_i$ such that 
$a_i=1$ if $p_i\in I$; otherwise, $a_i=0$. 
We write ${\sf row}_i(\bm{v})=p_i$. 
Given $\bm{v}=(a_1,\ldots,a_n)^{\sf T}\in$ {\Bb R}$^n$, define $\bm{v}[i]=a_i$  
$(1\leq i\leq n)$ and $\bm{v}[1\ldots k]=(a_1,\ldots,a_k)^{\sf T}\in$ {\Bb R}$^k$ $(k\leq n)$. 
\end{definition}

\begin{definition}[matrix representation of SD programs]\rm \label{p-matrix}
Let $P$ be an SD program and $B_P=\{ p_1,\ldots,p_n \}$. 
Then $P$ is represented by a matrix $\bm{M}_P\in$ {\Bb R}$^{n\times n}$ such that 
for each element $a_{ij}$ $(1\leq i,j\leq n)$ in $\bm{M}_P$, 
\begin{enumerate}
\item $a_{ij_k}=\frac{1}{m}\;\; (1\leq k\leq m;\, 1\leq i,j_k\leq n)$ if\, 
$p_i\leftarrow p_{j_1}\wedge\cdots\wedge p_{j_m}$ is in $P$;  
\item $a_{ii}=1$ if $p_i\leftarrow$ is in $P$; 
\item $a_{ij}=0$, otherwise.  
\end{enumerate}
$\bm{M}_P$ is called a {\em program matrix\/}. 
We write ${\sf row}_i(\bm{M}_P)=p_i$ and ${\sf col}_j(\bm{M}_P)=p_j$ $(1\leq i, j\leq n)$.%
\end{definition}

In $\bm{M}_P$ the $i$-th row corresponds to the atom $p_i$ 
appearing in the head of a rule, and the $j$-th column corresponds to the 
atom $p_j$ appearing in the body of a rule.  On the other hand, 
every fact $p_i\leftarrow$ in $P$ is represented as a tautology $p_i\leftarrow p_i$ in $\bm{M}_P$. 

\begin{example}\rm \label{ex-1}
Consider 
$P=\{\, p\leftarrow q,\;\;\; q\leftarrow p\wedge r,\;\;\;  r\leftarrow s,\;\;\; s\leftarrow \,\}$ 
with $B_P=\{\, p, q, r, s\,\}$. 
Then $\bm{M}_P$ becomes  
\begin{eqnarray*}
&& \qquad\qquad\quad\; p\;\;\;\;\;\; q \;\;\;\;\;\; r \;\;\;\;\;\;\; s\\
&& \qquad\begin{array}{c}   p \\ q\\ r\\ s\\ 
     \end{array}  
  \left ( \begin{array}{cccc} 
                                    
                                    0 & 1 & 0 & 0  \\
                                    \rfrac{1}{2} & 0 & \rfrac{1}{2} & 0  \\
                                    0 & 0 & 0 & 1  \\
                                    0 & 0 & 0 & 1   
                                  \end{array}  \right ) 
\end{eqnarray*} 
where ${\sf row}_1(\bm{M}_P)=p$ and ${\sf col}_2(\bm{M}_P)=q$. 
\end{example}

\begin{definition}[initial vector]\rm
Let $P$ be a definite program and $B_P=\{ p_1,\ldots,p_n \}$. 
Then the {\em initial vector\/} of $P$ is an interpretation vector 
$\bm{v}_0=(a_1,\ldots,a_n)^{\sf T}$ 
such that $a_i=1$ $(1\leq i\leq n)$ 
if ${\sf row}_i(\bm{v}_0)=p_i$ and a fact $p_i\leftarrow$ is in $P$; otherwise, $a_i=0$. 
\end{definition}

\begin{definition}[$\theta$-thresholding]\rm
Given a vector $\bm{v}=(a_1,\ldots,a_n)^{\sf T}$, define 
$\theta(\bm{v})=(a'_1,\ldots,a'_n)^{\sf T}$ where 
$a'_i=1$ $(1\leq i\leq n)$ if $a_i\geq 1$; otherwise, $a'_i=0$.%
\footnote{$a_i$ can be greater than 1 only later when d-rules come into play.}
We call $\theta(\bm{v})$ the\, $\theta$-{\em thresholding\/} of $\bm{v}$. 
\end{definition}

Given a program matrix $\bm{M}_P\in$ {\Bb R}$^{n\times n}$ and an initial vector 
$\bm{v}_0\in$ {\Bb R}$^{n}$, define 
\[ \bm{v}_1 = \theta(\bm{M}_P \bm{v}_0)\;\;\;\mbox{and}\;\;\;  
\bm{v}_{k+1} = \theta(\bm{M}_P \bm{v}_k)\;\;\;\; (k\geq 1) \]

It is shown that $\bm{v}_{k+1} = \bm{v}_k$ for some $k\geq 1$. 
When $\bm{v}_{k+1} = \bm{v}_k$, we write $\bm{v}_k = {\sf FP}(\bm{M}_P \bm{v}_0)$. 

\begin{theorem} \label{lm-th} 
Let $P$ be an SD program and $\bm{M}_P\in$ {\Bb R}$^{n\times n}$ its program matrix. 
Then $\bm{m}\in$ {\Bb R}$^n$ is a vector representing the least model of $P$ iff 
$\bm{m}={\sf FP}(\bm{M}_P \bm{v}_0)$ where $\bm{v}_0$ is the initial vector of $P$. 
\end{theorem}

\begin{example}\rm 
Consider the program $P$ of Example~\ref{ex-1} and its program matrix $\bm{M}_P$. 
The initial vector of $P$ is $\bm{v}_0 = (0,0,0,1)^{\sf T}$.  Then 
\begin{eqnarray*}
\bm{M}_P\bm{v}_0 = \left ( \begin{array}{cccc} 
                                    
                                    0 & 1 & 0 & 0  \\
                                    \rfrac{1}{2} & 0 & \rfrac{1}{2} & 0  \\
                                    0 & 0 & 0 & 1  \\
                                    0 & 0 & 0 & 1   
                                  \end{array}  \right ) 
  \left ( \begin{array}{cccc} 
                                    0  \\
                                    0  \\
                                    0  \\
                                    1   
                                  \end{array}  \right ) 
= \left ( \begin{array}{cccc} 
                                    
                                    0  \\
                                    0  \\
                                    1  \\
                                    1   
                                  \end{array}  \right ) 
\end{eqnarray*} 
and $\bm{v}_1 = \theta(\bm{M}_P \bm{v}_0)=(0,0,1,1)^{\sf T}$. Next, 
\begin{eqnarray*}
\bm{M}_P\bm{v}_1 =  \left ( \begin{array}{cccc}     
                                    0 & 1 & 0 & 0  \\
                                    \rfrac{1}{2} & 0 & \rfrac{1}{2} & 0  \\
                                    0 & 0 & 0 & 1  \\
                                    0 & 0 & 0 & 1   
                                  \end{array}  \right ) 
  \left ( \begin{array}{cccc} 
                                    0  \\
                                    0  \\
                                    1  \\
                                    1   
                                  \end{array}  \right ) 
= \left ( \begin{array}{cccc} 
                                    0  \\
                                    \rfrac{1}{2}  \\
                                    1  \\
                                    1   
                                  \end{array}  \right ) 
\end{eqnarray*} 
and $\bm{v}_2 = \theta(\bm{M}_P \bm{v}_1)=\bm{v}_1$.
Hence, $\bm{v}_2= (0,0,1,1)^{\sf T}$ represents the least model $\{r,s\}$ of $P$. 
\end{example}

\noindent {\em Remark:} 
The current study is different from the previous work \cite{SIS17} in matrix representation of 
programs as follows.  
\begin{itemize}
\item In \cite{SIS17} a fact is represented as a rule ``$p_i\leftarrow\top$" 
and is encoded in a matrix by $a_{ij}=1$ where ${\sf row}_i(\bm{M}_P)=p_i$ and 
${\sf col}_j(\bm{M}_P)=\top$. 
In contrast to the current study, the previous study 
sets the empty set as the initial vector and computes fixpoints.
In this study, we start with the initial vector representing facts, 
instead of representing facts as rules in $\bm{M}_P$.
This has the effect of increasing zero elements in matrices 
and reducing the number of required iterations in fixpoint computation. 
Representing matrices in sparse forms also 
brings storage advantages with a good matrix library.  

\item In \cite{SIS17} a constraint is represented as a rule 
``$\bot\leftarrow p_{j_1}\wedge\cdots\wedge p_{j_m}$" and is encoded in a matrix 
by $a_{ij_k}=\frac{1}{m}\;\; (1\leq k\leq m)$ where ${\sf row}_i(\bm{M}_P)=\bot$ and 
${\sf col}_{j_k}(\bm{M}_P)=p_{j_k}$. 
In the current study, we do not include constraints in a program as it causes a problem in 
partial evaluation.  Still, we can handle constraints separately from a program as follows.  
Given a program $P$ and constraints $C$, encode them by matrices 
$\bm{M}_P\in$ {\Bb R}$^{n\times n}$ and $\bm{M}_C\in $ {\Bb R}$^{(n+1)\times n}$, respectively, 
where $\bm{M}_C$ has the element $\bot$ in its row. 
After computing the fixpoint $\bm{v}_k = {\sf FP}(\bm{M}_P \bm{v}_0)\in$ {\Bb R}$^n$ as
in Theorem~\ref{lm-th}, 
compute $\bm{M}_C\bm{v}_k\in ${\Bb R}$^{n+1}$.  If ${\sf row}_i(\bm{v}_k)=\bot$ and 
$\bm{v}_k [i]=a_i=1$, then 
$P\cup C$ is inconsistent; otherwise, $\bm{v}_k$ represents the least model of $P\cup C$. 
\end{itemize}

\ignore{ 
\citeNS{SIS17} introduce fixpoint computation of least models in vector spaces. 
Different from the current study, the previous study 
sets the empty set as the initial vector and computes fixpoints.
In this study, we start by the initial vector representing facts, 
instead of representing facts in a matrix explicitly.
This has the effect of increasing 0s  
in matrices during fixpoint computation. 
Representing matrices in sparse forms  
brings storage advantages with a good matrix library.  
\cite{SIS17} allows the existence of constraints ``$\leftarrow G$" in a program, while the
current study does not handle constraints.%
\footnote{A constraint is represented as a rule ``$\bot\leftarrow G$" and is encoded in a matrix by 
preparing the element $\bot$ in the row and the column \cite{SIS17}. 
In the current study, we do not include constraints in a program as it causes a problem in 
partial evaluation.  Still we can handle constraints separately from a program as follows.  
Given a program $P$ and constraints $C$, encode them by matrices 
$\bm{M}_P\in$ {\Bb R}$^{n\times n}$ and $\bm{M}_C\in $ {\Bb R}$^{(n+1)\times n}$, respectively, 
where $\bm{M}_C$ has the element $\bot$ in its row. 
After computing the fixpoint $\bm{v}_k = {\sf FP}(\bm{M}_P \bm{v}_0)\in$ {\Bb R}$^n$,  
compute $\bm{M}_C\bm{v}_k\in ${\Bb R}$^{n+1}$.  If ${\sf row}_i(\bm{v})=\bot$ and $a_i=1$, then 
$P\cup C$ is inconsistent; otherwise, $\bm{v}_k$ is the least model of $P\cup C$. } 
}

\subsection{Non-SD programs}  \label{sec:3.2}

When a definite program $P$ contains two rules: 
$r_1:\, h\leftarrow b_1\wedge\cdots\wedge b_m$ and $r_2:\, h\leftarrow c_1\wedge\cdots\wedge c_n$, 
$P$ is transformed to a d-program $P^{\delta}=(P\setminus \{ r_1, r_2\})\cup \{ r'_1, r'_2, d_1 \}$ 
where $r'_1:\, h_1\leftarrow b_1\wedge\cdots\wedge b_m$, $r'_2:\, h_2\leftarrow c_1\wedge\cdots\wedge c_n$ and $d_1:\, h\leftarrow h_1\vee h_2$. 
Here, $h_1$ and $h_2$ are new propositional variables associated with $r_1$ and $r_2$, respectively. 

Generally, a non-SD program is transformed to a d-program as follows. 

\begin{definition}[transformation]\rm
Let $P$ be a definite program and $B_P$ its Herbrand base. 
For each $p\in B_P$, put $P_p=\{\, r \mid r\in P\;\mbox{and}\; head(r)=p\,\}$ and 
$R_p=\{\, r \mid r\in P_p\;\mbox{and}\, \mid\! P_p\!\mid\, >1 \,\}$. 
Then define  
$S_p=\{\, p_i\leftarrow body(r) \mid  r\in R_p\;\mbox{and}\; i=1,\ldots,k\;\mbox{where}\; k=\mid \! R_p\!\mid\,\}$ 
and $D_p=\{\, p\leftarrow p_1\vee\cdots\vee p_k\,\mid\, p_i\leftarrow body(r)\;\mbox{is in}\; S_p\}$ 
where $p_i$ is a new propositional variable such that $p_i\not\in B_P$ and $p_i\neq p_j$ if $i\neq j$. 
Then, build a d-program 
\[ P^{\delta} = (P \setminus \bigcup_{p\in B_P} R_p) \cup \bigcup_{p\in B_P} (S_p \cup D_p)\]
where $Q = (P \setminus \bigcup_{p\in B_P} R_p) \cup \bigcup_{p\in B_P} S_p$ is an SD program 
and $D=\bigcup_{p\in B_P} D_p$ is a set of d-rules. 
\end{definition}

$P^{\delta}$ introduces additional propositional variables and $B_P\subseteq B_{P^{\delta}}$ holds. 
By definition, the next result holds. 

\begin{proposition}  \label{prop-pos} 
Let $P$ be a definite program and $P^{\delta}$ its transformed d-program. 
Suppose that $P$ and $P^{\delta}$ have the least models $M$ and $M'$, respectively. 
Then $M=M'\cap B_P$ holds. 
\end{proposition}

In this way, any definite program $P$ is transformed to a semantically equivalent 
d-program $P^{\delta}=Q\cup D$ where $Q$ is an SD program and $D$ is a set of d-rules. 
A d-program is represented by a matrix as follows. 

\begin{definition}[program matrix for d-programs]\rm \label{p-matrix2}
Let $P^{\delta}$ be a d-program such that $P^{\delta}=Q\cup D$ 
where $Q$ is an SD program and $D$ is a set of d-rules, 
and $B_{P^{\delta}}=\{ p_1,\ldots,p_m \}$ the Herbrand base of $P^{\delta}$. 
Then $P^{\delta}$ is represented by a matrix $\bm{M}_{P^{\delta}}\in$ {\Bb R}$^{m\times m}$ such that 
for each element $a_{ij}$ $(1\leq i,j\leq m)$ in $\bm{M}_{P^{\delta}}$, 
\begin{enumerate}
\item $a_{ij_k}=1\;\; (1\leq k\leq l;\, 1\leq i,j_k\leq m)$ if\, 
$p_i\leftarrow p_{j_1}\vee \cdots \vee p_{j_l}$ is in $D$;  
\item otherwise, every rule in $Q$ is encoded as in Def.~\ref{p-matrix}. 
\end{enumerate}
\end{definition}

Given a program matrix $\bm{M}_{P^{\delta}}$ and the initial vector $\bm{v}_0$ representing facts 
in $P^{\delta}$, the fixpoint $\bm{v}_k={\sf FP}(\bm{M}_{P^{\delta}}\bm{v}_0)$ $(k\geq 1)$ 
is computed as before.  
The fixpoint represents the least model of $P^{\delta}$. 

\begin{theorem} \label{fp-pos} 
Let $P^{\delta}$ be a d-program and $\bm{M}_{P^{\delta}}\in$ {\Bb R}$^{m\times m}$ its program matrix. 
Then $\bm{m}\in$ {\Bb R}$^m$ is a vector representing the least model of $P^{\delta}$ iff 
$\bm{m}={\sf FP}(\bm{M}_{P^{\delta}} \bm{v}_0)$ where $\bm{v}_0$ is the initial vector of $P^{\delta}$. 
\end{theorem}

By Proposition~\ref{prop-pos} and Theorem~\ref{fp-pos}, we can compute the least model of 
any definite program.  

\begin{example}\rm \label{ex-dprogram}
Consider the program
$P=\{\, p\leftarrow q,\;\;\; q\leftarrow p\wedge r,\;\;\;  q\leftarrow s,\;\;\; s\leftarrow \,\}$. 
As $P$ is a non-SD program,\\ 
\begin{minipage}[l]{7.2cm} 
it is transformed to a d-program 
$P^{\delta}=\{\, p\leftarrow~q,\\ t\leftarrow p\wedge r,\;\;  u\leftarrow s,\;\; s\leftarrow,\;\; q\leftarrow t\vee u \,\}$ where $t$ and $u$ are new propositional variables.  
Then $\bm{M}_{P^{\delta}}\in$ {\Bb R}$^{6\times 6}$ becomes the matrix (right). 
The initial vector of $P^{\delta}$ is $\bm{v}_0 = (0,0,0,1,0,0)^{\sf T}$.  Then,  
$\bm{v}_1 = \theta(\bm{M}_{P^{\delta}} \bm{v}_0)=(0,0,0,1,0,1)^{\sf T}$, 
$\bm{v}_2 = \theta(\bm{M}_{P^{\delta}} \bm{v}_1)=(0,1,0,1,0,1)^{\sf T}$, 
$\bm{v}_3 = \theta(\bm{M}_{P^{\delta}} \bm{v}_2)=(1,1,0,1,0,1)^{\sf T}$, and 
$\bm{v}_4 = \theta(\bm{M}_{P^{\delta}} \bm{v}_3)= \bm{v}_3$. 
Then  $\bm{m}={\sf FP}(\bm{M}_{P^{\delta}} \bm{v}_0)=(1,1,0,1,0,1)^{\sf T}$ represents the least model 
$\{p,q,s,u\}$ of $P^{\delta}$, hence $\{p,q,s,u\}\cap B_P=\{ p,q,s\}$ is the least model of $P$. 
\end{minipage}
\begin{minipage}[r]{5cm}
\vspace*{-1cm}
\begin{eqnarray*}
&& \qquad\quad\; p\;\;\;\;\;\; q \;\;\;\;\;\;\, r \;\;\;\;\;\, s\;\;\;\;\;\; t\;\;\;\;\;\; u\\
&& \begin{array}{c}   p \\ q\\ r\\ s\\ t\\ u 
     \end{array}  
  \left ( \begin{array}{cccccc} 
                                    
                                    0 & 1 & 0 & 0 & 0 & 0\\
                                    0 & 0 & 0 & 0 & 1 & 1 \\
                                    0 & 0 & 0 & 0 & 0 & 0 \\
                                    0 & 0 & 0 & 1 & 0 & 0 \\
                                    \rfrac{1}{2} & 0 & \rfrac{1}{2} & 0 & 0 & 0 \\
                                    0 & 0 & 0 & 1 & 0 & 0 
                                  \end{array}  \right ) 
\end{eqnarray*} 
\end{minipage}
\end{example}

\begin{figure}[t]
\hrulefill \newline
{\bf Algorithm~1: Matrix Computation of Least Models} 
\begin{description}	
\item[Input:]  a definite program $P$ and its Herband base $B_P = \{p_1,\ldots,p_n\}$.
\item[Output:] a vector $\bm{u}$ representing the least model of $P$. 
\item[] {\bf Step 1:} Transform $P$ to a d-program $P^{\delta}=Q\cup D$ with 
$B_{P^{\delta}}=\{\, p_1,\ldots, p_n, p_{n+1}, \ldots, p_m \,\}$ where $Q$ is an 
SD program and $D$ is a set of d-rules. 
\item[] {\bf Step 2:} Embed $P^{\delta}$ into a vector space. 
\item[] \hspace*{7mm} - Create the matrix $\bm{M}_{P^{\delta}}\in$ {\Bb R}$^{m\times m}$  representing $P^{\delta}$. 
\item[] \hspace*{7mm} - Create the initial vector $\bm{v}_0=(v_1,\ldots,v_m)^{\sf T}$ of $P^{\delta}$. 
\item[] {\bf Step 3:}  Compute the least model of $P^{\delta}$. 
\item[] \hspace*{10mm} $\bm{v}:= \bm{v}_0$;  
\item[] \hspace*{10mm} $\bm{u}:= \theta(\bm{M}_{P^{\delta}}\bm{v})$
\item[] \hspace*{10mm} while $\bm{u}\neq \bm{v}$ do  
\item[] \hspace*{15mm} $\bm{v}:= \bm{u}$; 
\item[] \hspace*{15mm} $\bm{u}:= \theta(\bm{M}_{P^{\delta}}\bm{v})$ 
\item[] \hspace*{10mm} end do					
\item[] \hspace*{7mm} return $\bm{u}[1\ldots n]$
\item[]\hrulefill
\end{description}
\caption{Algorithm for computing least models} \label{algo-1}
\end{figure}

An algorithm for computing the least model of a definite program $P$ is shown in Figure~\ref{algo-1}. 
In the algorithm, the complexity of computing ${\bm M}_{P^{\delta}}{\bm v}$ is $O(m^2)$ and 
computing $\theta(\cdot)$ is $O(m)$. 
The number of times for iterating ${\bm M}_{P^{\delta}}{\bm v}$ is at most $(m + 1)$ times. 
So the complexity of Step 3 is $O((m + 1)\times (m+m^2))=O(m^3)$ in the worst case.

\subsection{Column Reduction}  \label{sec:3.3}

To decrease the complexity of computing ${\bm M}_{P^{\delta}}{\bm v}$, we introduce a technique of 
column reduction of program matrices. 

\begin{definition}[submatrix representation of d-programs]\rm \label{def-submatrix}
Let $P$ be a definite program such that $\mid\! B_P\mid=n$. 
Suppose that $P$ is transformed to a d-program 
$P^{\delta}$ such that $P^{\delta}=Q\cup D$ where $Q$ is an SD program 
and $D$ is a set of d-rules, and $B_{P^{\delta}}=\{p_1,\ldots, p_m\}$. 
Then $P^{\delta}$ is represented by a matrix $\bm{N}_{P^{\delta}}\in$ {\Bb R}$^{m\times n}$
such that each element $b_{ij}$ $(1\leq i\leq m;\, 1\leq j\leq n)$ in $\bm{N}_{P^{\delta}}$ 
is equivalent to the corresponding element $a_{ij}$ $(1\leq i, j \leq m)$ in $\bm{M}_{P^{\delta}}$ of Def.~\ref{p-matrix2}.  $\bm{N}_{P^{\delta}}$ is called a {\em submatrix\/} of $P^{\delta}$.
\end{definition}

Note that the size of $\bm{M}_{P^{\delta}}\in$ {\Bb R}$^{m\times m}$ of Def.~\ref{p-matrix2} is 
reduced to $\bm{N}_{P^{\delta}}\in$ {\Bb R}$^{m\times n}$ in Def.~\ref{def-submatrix} by 
$n\leq m$.  In $\bm{N}_{P^{\delta}}$ the columns do not include values of newly introduced 
propositions and derivation of propositions in $B_P$ via 
d-rules is checked by the following $\theta_D$-thresholding. 

\begin{definition}[$\theta_D$-thresholding]\rm \label{df-dthres}
Given a vector $\bm{v}=(a_1,\ldots,a_m)^{\sf T}$,  
define a vector $\bm{w}=\theta_D(\bm{v})=(w_1,\ldots,w_m)^{\sf T}$ such that 
(i) $w_i=1$ $(1\leq i\leq m)$ if $a_i\geq 1$, 
(ii) $w_i=1$ $(1\leq i\leq n)$ if $\exists j$ $w_j=1$ $(n+1\leq j\leq m)$ and there is a 
d-rule $d\in D$ such that $head(d)=p_i$ and ${\sc row}_j(\bm{w})\in body(d)$, and (iii) 
otherwise, $w_j=0$. We call $\theta_D(\bm{v})$ the\, $\theta_D$-{\em thresholding\/} of $\bm{v}$. 
\end{definition}

Intuitively, $\theta_D$-thresholding introduces an additional condition Def.~\ref{df-dthres}(ii) to $\theta$-thresholding, 
which means that 
``if an element in the body of a d-rule is 1, then the element in the head of the d-rule is set to 1''.  
$\theta_D(\bm{v})$ is computed by checking the value of $a_i$ for  $1\leq i\leq m$ and 
checking all d-rules for $n+~1\leq j\leq m$.  
Since the number of d-rules is at most $n$, 
the complexity of computing $\theta_D(\bm{v})$ is $O(m + (m-n)\times n)=O(m\times n)$.  
By definition, it holds that $\theta_D(\bm{v})=\theta_D(\theta(\bm{v}))$. 

\begin{proposition}  \label{subm-prop}
Let $P$ be a definite program with $B_P=\{ p_1,\ldots, p_n \}$, and $P^{\delta}$ a transformed 
d-program with $B_{P^{\delta}}=\{ p_1,\ldots, p_n, p_{n+1},\ldots, p_m \}$. 
Let $\bm{N}_{P^{\delta}}\in$ {\Bb R}$^{m\times n}$ be a submatrix of $P^{\delta}$. 
Given a vector $\bm{v}\in$ {\Bb R}$^n$ representing an interpretation $I$ of $P$, 
let $\bm{u}=\theta_D(\bm{N}_{P^{\delta}} \bm{v})\in$ {\Bb R}$^m$. 
Then $\bm{u}$ is a vector representing an interpretation $J$ of $P^{\delta}$ such that 
$J\cap B_P=T_P(I)$. 
\end{proposition}

Given a program matrix $\bm{N}_{P^{\delta}}\in$ {\Bb R}$^{m\times n}$ and the initial vector 
$\bm{v}_0\in$ {\Bb R}$^{m}$ of $P^{\delta}$, define 
\[ \bm{v}_1 = \theta_D(\bm{N}_{P^{\delta}} \bm{v}_0[1\ldots n])\;\;\;\mbox{and}\;\;\;  
\bm{v}_{k+1} = \theta_D(\bm{N}_{P^{\delta}}\bm{v}_k[1\ldots n])\;\;\;\; (k\geq 1) \]
where $\bm{N}_{P^{\delta}}\bm{v}_k[1\ldots n]$ represents the product of 
$\bm{N}_{P^{\delta}}$ and $\bm{v}_k[1\ldots n]$. 
Then it is shown that $\bm{v}_{k+1}=\bm{v}_{k}$ for some $k\geq 1$. 
When $\bm{v}_{k+1}=\bm{v}_{k}$, we write $v_k={\sf FP}(\bm{N}_{P^{\delta}}\bm{v}_0[1\ldots n])$. 

\begin{theorem}
Let $P$ be a definite program with $B_P=\{ p_1,\ldots, p_n \}$, and $P^{\delta}$ a transformed 
d-program with $B_{P^{\delta}}=\{ p_1,\ldots, p_n, p_{n+1},\ldots, p_m \}$. 
Then $\bm{m}\in$ {\Bb R}$^n$ is a vector representing the least model of $P$ iff 
$\bm{m}={\sf FP}(\bm{N}_{P^{\delta}}\bm{v}_0[1\ldots n])$ 
where $\bm{v}_0\in$ {\Bb R}$^m$ is the initial vector of $P^{\delta}$. 
\end{theorem}

Generally, given a d-program $P^{\delta}$, the value $k$ of 
$\bm{v}_k={\sf FP}(\bm{N}_{P^{\delta}}\bm{v}_0[1\ldots n])$ is not greater than the value $h$ of 
$\bm{v}_h={\sf FP}(\bm{M}_{P}\bm{v}_0)$ of Section~\ref{sec:3.1}. 

\begin{example}
For the d-program $P^{\delta}$ of Example~\ref{ex-dprogram}, 
we have the submatrix $\bm{N}_{P^{\delta}}\in$ {\Bb R}$^{6\times 4}$ representing $P^{\delta}$. \\
\begin{minipage}[l]{8cm} 
Given the initial vector $\bm{v}_0=(0,0,0,1,0,0)^{\sf T}$ of $P^{\delta}$, it becomes 
$\bm{v}_1 = \theta_D(\bm{N}_{P^{\delta}} \bm{v}_0[1\ldots 4])=(0,1,0,1,0,1)^{\sf T}$, 
$\bm{v}_2 = \theta_D(\bm{N}_{P^{\delta}} \bm{v}_1[1\ldots 4])=(1,1,0,1,0,1)^{\sf T}$, 
$\bm{v}_3 = \theta_D(\bm{N}_{P^{\delta}} \bm{v}_2[1\ldots 4])=(1,1,0,1,0,1)^{\sf T}=\bm{v}_2$.
Then $\bm{v}_2$ is a vector representing the least model of $P^{\delta}$, and 
$\bm{v}_2[1\ldots 4]$ is a vector representing the least model $\{p,q,s\}$ of $P$. 
Note that the second element of $\bm{v}_i$ $(i=1,2,3)$ becomes $1$ by Def.~\ref{df-dthres}(ii). 
\end{minipage}
\begin{minipage}[r]{4cm}
\vspace*{-1cm}
\begin{eqnarray*}
&& \qquad\quad\; p\;\;\;\;\;\, q \;\;\;\;\;\, r \;\;\;\;\;\;\, s\\
&& \begin{array}{c}   p \\ q\\ r\\ s\\ t\\ u 
     \end{array}  
  \left ( \begin{array}{cccc} 
                                    
                                    0 & 1 & 0 & 0\\
                                    0 & 0 & 0 & 0\\
                                    0 & 0 & 0 & 0\\
                                    0 & 0 & 0 & 1\\
                                    \rfrac{1}{2} & 0 & \rfrac{1}{2} & 0\\
                                    0 & 0 & 0 & 1 
            \end{array}  \right ) 
\end{eqnarray*} 
\end{minipage}
\end{example}

By Proposition~\ref{subm-prop}, we can replace the computation 
$\bm{u}=\theta(\bm{M}_{P^{\delta}}\bm{v})$ in Step~3 of Algorithm~1 in Figure~\ref{algo-1} 
by $\bm{u}=\theta_D(\bm{N}_{P^{\delta}}\bm{v}[1\ldots n])$. 
In the column reduction method, the complexity of computing $\bm{N}_{P^{\delta}}\bm{v}$ is 
$O(m\times n)$ and computing $\theta_D(\cdot)$ is $O(m\times n)$. 
The number of times for iterating $\bm{N}_{P^{\delta}}\bm{v}$ is at most $(m+1)$ times. 
So the complexity of computing $\bm{u}=\theta_D(\bm{N}_{P^{\delta}}\bm{v}[1\ldots n])$ is 
$O((m+1)\times (m\times n + m\times n))=O(m^2\times n)$. 
Comparing the complexity $O(m^3)$ of Step~3 in Algorithm~1, 
the column reduction reduces the complexity to $O(m^2\times n)$ as $m\gg n$ in general. 


\section{Partial Evaluation}  \label{sec:4}

{\em Partial evaluation\/} is known as an optimization technique in logic programming \cite{LS91}.  
In this section, we provide a method of computing partial evaluation of definite programs in 
vector spaces. 

\begin{definition}[partial evaluation]\rm \label{df-peval}
Let $P$ be an SD program. 
For any rule $r$ in $P$, put $U_r=\{\,r_i\,\mid\, r_i\in P\;\mbox{and}\; head(r_i)\in body(r)\,\}$. 
Then construct a rule $r'={\sf unfold}(r)$ such that 
\begin{itemize}
\item $head(r')=head(r)$, and 
\item $body(r')=(body(r)\setminus \bigcup_{r_i\in U_r}\,\{head(r_i)\})\,\cup \bigcup_{r_i\in U_r}\,body(r_i).$
\end{itemize}
Define 
\[ {\sf peval}(P)=(\,\bigcup_{r\in P}\, {\sf unfold}(r)\,)\, \setminus R\] 
where $R=\{\, r \mid body(r)\cap (B_P\setminus H_P) \neq \emptyset\,\}$ and 
$H_P=\{\, a \mid \mbox{there is a rule $r$ in $P$ s.t. $head(r)=a$}\,\}$. 
${\sf peval}(P)$ is called {\em partial evaluation\/} of $P$. 
\end{definition}

\begin{example} \rm \label{ex-peval}
Consider 
$P=\{\, p\leftarrow~q\wedge s\wedge t,\;\;\; q\leftarrow~p\wedge t,\;\;\; s\leftarrow t,\;\;\; t\leftarrow\,\}$. 
Put $r_1=(p\leftarrow~q\wedge s\wedge t)$, $r_2=(q\leftarrow~p\wedge t)$, 
$r_3=(s\leftarrow t)$, and $r_4=(t\leftarrow)$.  Unfolding rules produces: 
${\sf unfold}(r_1)=(p\leftarrow p\wedge t\wedge t)=(p\leftarrow p\wedge t)$, 
${\sf unfold}(r_2)=(q\leftarrow q\wedge s\wedge t)$, 
${\sf unfold}(r_3)=(s\leftarrow)$, and 
${\sf unfold}(r_4)=(t\leftarrow)$. Then it becomes 
${\sf peval}(P)=\{\, p\leftarrow p\wedge t,\;\; q\leftarrow q\wedge s\wedge t,\;\; s\leftarrow,\;\; t\leftarrow\,\}$.  
\end{example}

By definition, ${\sf peval}(P)$ is obtained from $P$ by unfolding propositional variables 
appearing in the body of any rule in $P$ in parallel.  
If $body(r)$ contains an atom unfolded by no rule in $P$, then $r$ is just removed from $P$. 
Partial evaluation preserves the least model of the original program \cite{LS91}. 

\begin{proposition}
Let $P$ be an SD program. 
Then $P$ and ${\sf peval}(P)$ have the same least model.  
\end{proposition}

Partial evaluation is computed by matrix products in vector spaces. 

\begin{example}\rm
The program $P$ of Example~\ref{ex-peval} is represented by the matrix $\bm{M}_P$, and 
$(\bm{M}_P)^2$ becomes 
\begin{eqnarray*}
&& \qquad\qquad\qquad\; p\;\;\;\;\;\, q \;\;\;\;\;\;\;\; s \;\;\;\;\;\;\;\;\; t\\
&& \bm{M}_P =\begin{array}{c}   p \\ q\\ s\\ t
     \end{array}  
  \left ( \begin{array}{cccc} 
                                    
                                    0 & \rfrac{1}{3} & \rfrac{1}{3} & \rfrac{1}{3}   \\
                                    \rfrac{1}{2} & 0 & 0 & \rfrac{1}{2}   \\
                                    0 & 0 & 0 & 1  \\
                                    0 & 0 & 0 & 1    
                                  \end{array}  \right ) 
\qquad (\bm{M}_P)^2 =  \left ( \begin{array}{cccc} 
                                    
                                    \rfrac{1}{6} & 0 & 0 & \rfrac{5}{6}  \\
                                    0 & \rfrac{1}{6} & \rfrac{1}{6} & \rfrac{2}{3}  \\
                                    0 & 0 & 0 & 1  \\
                                    0 & 0 & 0 & 1  
                                  \end{array}  \right ) 
\end{eqnarray*} 
\end{example} 

Intuitively speaking, non-zero elements in $(\bm{M}_P)^2$ represent conjuncts appearing in each rule. 
So the first row represents the rule $p\leftarrow p\wedge t$ and the second row 
represents the rule $q\leftarrow q\wedge s\wedge t$. $(\bm{M}_P)^2$ then represents 
$P'=\{\, p\leftarrow p\wedge t,\;\; q\leftarrow q\wedge s\wedge t,\;\; s\leftarrow t,\;\; t\leftarrow\,\}$. 
$P'$ is different from ${\sf peval}(P)$ for the representation of the rule $s\leftarrow t$. 
This is because $t\leftarrow$ is represented as $t\leftarrow t$ in $\bm{M}_P$, so that 
unfolding $s\leftarrow t$ by $t\leftarrow t$ becomes $s\leftarrow t$. 
Thus, $(\bm{M}_P)^2$ does not represent the result of unfolding rules by facts precisely, while 
it does not affect the result of computing the least model of $P$.  
In fact, applying the vector $\bm{v}_0=(0,0,0,1)^{\sf T}$ representing facts in $P$ and applying 
$\theta$-thresholding, we obtain $\theta((\bm{M}_P)^2 \bm{v}_0) = (0,0,1,1)$ 
that represents the least model $\{\, s, t \,\}$ of $P$. 
We say that $(\bm{M}_P)^2$ represents the {\em rule by rule\/} (shortly, {\em r-r\/}) partial evaluation, and often say just partial evaluation when no confusion arises. 
Formally, we have the next result. 

\ignore{ 
\begin{definition}[adjustment]\rm
Given a matrix $\bm{M}\in$ {\Bb R}$^{n\times n}$, 
the matrix $\gamma\bm{M}$ is obtained by replacing 
$a_{ij}(\neq 0)$ $(1\leq i,j\leq n)$ with $\frac{1}{k}$ 
when there are $k\,(>0)$ non-zero elements in the $i$-th row. 
\end{definition}

Applying $\gamma$ to the above example, it becomes 
\[
 \gamma(\bm{M}_P)^2 =  \left ( \begin{array}{cccc} 
                                    
                                    \rfrac{1}{2} & 0 & 0 & \rfrac{1}{2}  \\
                                    0 & \rfrac{1}{3} & \rfrac{1}{3} & \rfrac{1}{3}  \\
                                    0 & 0 & 0 & 1  \\
                                    0 & 0 & 0 & 1 
                                    \end{array}  \right ) 
\] 

Put $\Gamma_P=\gamma(\bm{M}_P)^2$. 
Then it becomes $\gamma(\Gamma_P)^2=\Gamma_P$. 
This means that $\Gamma_P$ is the fixpoint that represents the program 
$\{\, p\leftarrow p,t,\;\; q\leftarrow q,s,t,\;\; s\leftarrow t,\;\; t\leftarrow\,\}$ 
which is ${\sf peval}(P)$. 
}  

\begin{proposition} \label{peval-prop}
Let $P$ be an SD program and $\bm{v}_0$ the initial vector representing facts of $P$. 
Then $\theta((\bm{M}_P)^2\bm{v}_0)=\theta(\bm{M}_P(\theta(\bm{M}_P\bm{v}_0)))$.  
\end{proposition}

Partial evaluation has the effect of reducing deduction steps by unfolding rules in advance.  
Proposition~\ref{peval-prop} realizes this effect by computing matrix products in advance. 
Partial evaluation is performed iteratively as 
\[  {\sf peval}^k(P)={\sf peval}({\sf peval}^{k-1}(P))\;\; (k\geq 1)\;\;\;\mbox{and}\;\;\; {\sf peval}^0(P)=P. \]
Iterative partial evaluation is computed by matrix products as follows. 

Let $P$ be an SD program and $\bm{M}_P\in$ {\Bb R}$^{n\times n}$ its program matrix. 
Define $\Gamma^1_{P}=(\bm{M}_{P})^2$ and 
$\Gamma^{k+1}_{P}=(\Gamma^k_{P})^2$ $(k\geq 1)$.  
Then $\Gamma^k_{P}$ is a matrix representing a program that is obtained by 
$k$-th iteration of (r-r) partial evaluation. 

\begin{theorem} \label{peval-th}
Let $P$ be an SD program and $\Gamma^k_{P}\in$ {\Bb R}$^{n\times n}$ $(k\geq 1)$. 
Then $\theta(\Gamma^k_{P}\bm{v}_0)=\bm{v}_{2^k}$ 
where $\bm{v}_k=\theta(\bm{M}_P\bm{v}_{k-1})$.  
\end{theorem}

When $P$ is a non-SD program, first transform $P$ to a d-program 
$P^{\delta}=Q\cup D$ where $Q$ is an SD program and $D$ is a set of d-rules (Section~\ref{sec:3.2}). 
Next, define $\Gamma^k_{P^{\delta}} = \Gamma^k_{Q} + \bm{M}_D$. 
We then compute (r-r) partial evaluation of $P^{\delta}$ as (r-r) partial evaluation of an SD program $Q$ 
plus d-rules $D$. 


An algorithm for computing the least model of a definite program $P$ by (r-r) partial evaluation is shown in Figure~\ref{algo-2}.  We can combine partial evaluation and column reduction 
of Section~\ref{sec:3.3} by slightly changing Step~3 of Algorithm~2.  
We evaluate this hybrid method in the next section.  

\begin{figure}[t]
\hrulefill \newline
{\bf Algorithm~2: Partial Evaluation} 
\begin{description}	
\item[Input:]  a definite program $P$ and its Herband base $B_P$.\\
\hspace*{5mm}  $k\,(\geq 0)$:  the number of iteration of partial evaluation.  
\item[Output:] a vector $\bm{u}$ representing the least model of $P$. 
\item[] {\bf Step 1:} Transform $P$ to a d-program $P^{\delta}=Q\cup D$ 
where $Q$ is an SD program and $D$ is a set of d-rules. 
\item[] {\bf Step 2:} Embed $P^{\delta}$ into a vector space. 
\item[] \hspace*{7mm} - Create the matrix $\bm{M}_Q$ representing $Q$.  
\item[] \hspace*{7mm} - Create the matrix $\bm{M}_D$ representing $D$. 
\item[] {\bf Step 3:}  Compute (r-r) partial evaluation of $Q$. 
\item[] \hspace*{10mm} $\Gamma_Q^1=(\bm{M}_Q)^2$; 
\item[] \hspace*{10mm} $i:=1$; while $i\leq k$ do  
\item[] \hspace*{15mm} $\Gamma_Q^{i+1}=(\Gamma_Q^i)^2$; 
\item[] \hspace*{10mm} $i:=i+1$;  end do
\item[] \hspace*{10mm} Compute $\Gamma^k_{P^{\delta}} := \Gamma^k_{Q} + \bm{M}_D$
\item[] \hspace*{7mm} Create the vector $\bm{v}_0$ representing the facts of $Q$  
\item[] \hspace*{10mm} $\bm{v}:= \bm{v}_0$; 
\item[] \hspace*{10mm} $\bm{u}:= \theta(\Gamma^k_{P^{\delta}} \bm{v}_0)$; 
\item[] \hspace*{10mm} while $\bm{u}\neq \bm{v}$ do
\item[] \hspace*{15mm} $\bm{v}:= \bm{u}$; 
\item[] \hspace*{15mm} $\bm{u}:= \theta(\Gamma^k_{P^{\delta}} \bm{v})$; 
\item[] \hspace*{10mm} end do					
\item[] \hspace*{7mm} return $\bm{u}$
\item[]\hrulefill
\end{description}
\caption{Algorithm for computing least models by partial evaluation} \label{algo-2}
\end{figure}

\section{Experimental Results}

In this section, we compare runtime for computing the least model of a definite program. 
The testing is done on a computer with the following configuration: 
\begin{itemize}
\item Operating system: Linux Ubuntu 16.04 LTS 64bit
\item CPU: Intel Core$^{TM}$ i7-6800K (3.4 GHz/14nm/Cores=6/Threads=12/Cache=15MB), Memory 32GB, DDR-2400 
\item	GPU: GeForce GTX1070TI GDDR5 8GB
\item	Implementation language: Maple 2017, 64 bit
\footnote{Maple 2017 supports to use GPU for accelerating linear algebraic computation by 
CUDA package \cite{maple}.  The experimental results in this section do not use the CUDA 
package, however.} 
\end{itemize}


Given the size $n = \mid B_P\mid$ of the Herband base $B_P$ 
and the number $m = \mid P\mid$ of rules in $P$, rules are randomly generated as in Table~\ref{rule-rate}.  

\begin{table}[h]
\caption{Proportion of rules in $P$ based on the number of propositional variables in their bodies}
\label{rule-rate}
\begin{tabular}{cccccccccc}\hline\hline
Number of elements in body & 0 & 1 & 2 & 3 & 4 & 5 & 6 & 7 & 8 \\
Number of rules (proportion) 	& $x < \frac{n}{3}$ & 4\% &	4\% & 10\%	& 40\% & 35\% & 4\% & 2\% & 0-1\% \\
\hline\hline
\end{tabular}
\end{table}

A definite program $P$ is randomly generated based on $(n, m)$. 
We set those parameters as $n\ll m$, so generated programs are non-SD programs and they are transformed to d-programs. 
We compare runtime for computing the least model of $P$ by the following four methods: 
(a) computation by the $T_P$-operator;  
(b) computation by program matrices; (c) computation by column reduction; and  
(d) partial evaluation. 
\begin{figure}[t]
\hrulefill 
\begin{description}	
\item[Input:]  a definite program $P$.
\item[Output:] the least model of $P$. 
\item[] $I:=$ set of facts in $P$; 
\item[] $J:=\emptyset$; 
\item[]  while $(I\neq J)$ do
\item[] \hspace*{7mm} $J:=I$; 
\item[] \hspace*{7mm} for $r$ in $P$ do 
\item[] \hspace*{10mm} if $body(r)\subseteq J$ then $I:=I\cup \{ head(r)\}$;  
\item[] \hspace*{10mm} end do					
\item[] \hspace*{7mm} return $J$
\item[]\hrulefill
\end{description}
\caption{Algorithm for computing least models by $T_P$-operator} \label{algo-Tp}
\end{figure}
Computation by the $T_P$-operator is done by the procedure in Figure~\ref{algo-Tp}. 
Computation by program matrices is done by Algorithm~1, and computation by column reduction 
is done by modifying Step~3 of Algorithm~1 (see Sec.~\ref{sec:3.3}). 
In partial evaluation, the input parameter $k$ of Algorithm~2 is set as $k=1,5,\frac{n}{2}, n$ where 
$n = \mid B_P\mid$.  We then compute a vector representing the least model of $P$ in two ways:  program matrices and column reduction.  

\begin{figure}[h] 
\centering 
\includegraphics[width=90mm]{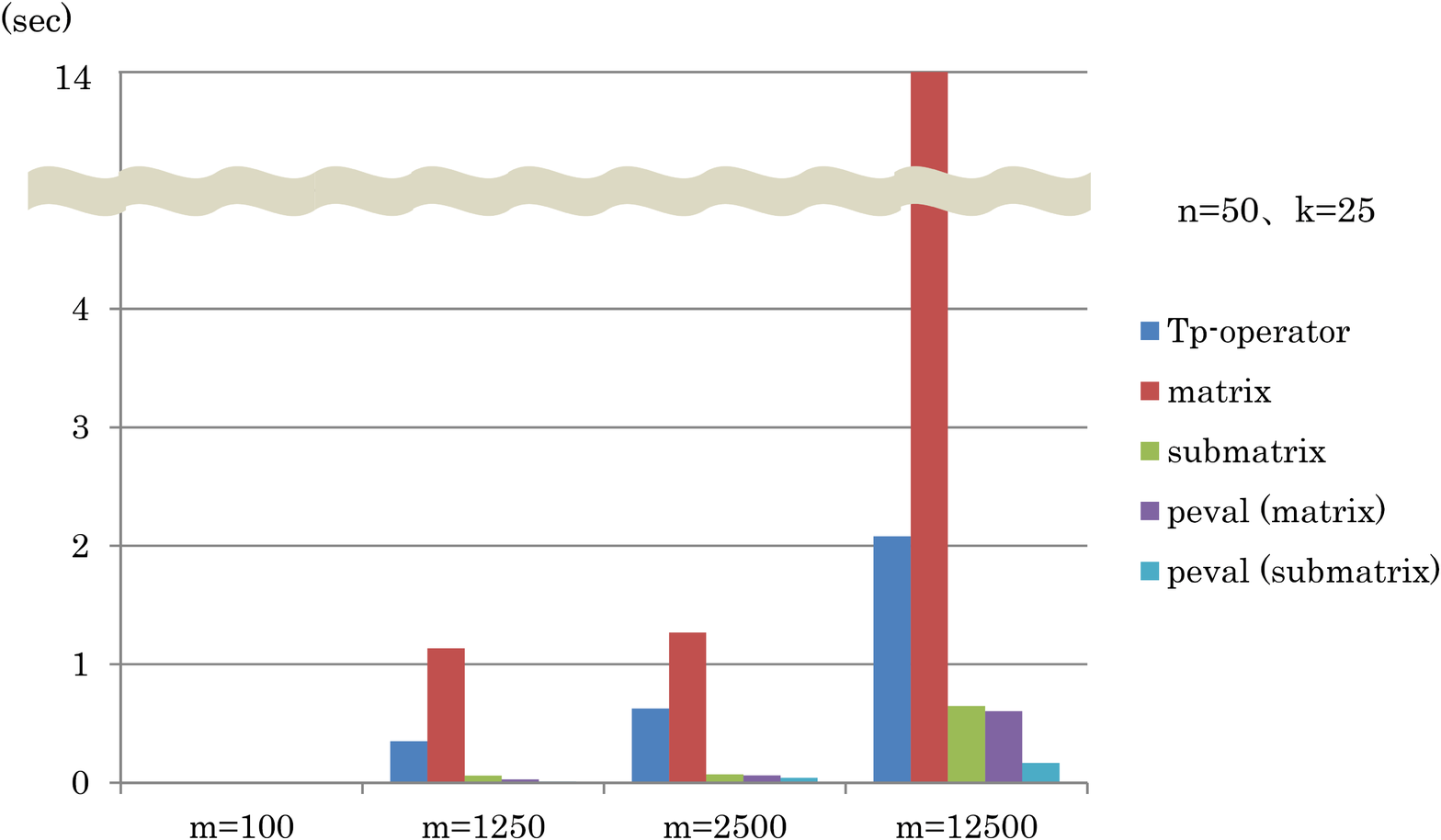} \\
\vspace*{10mm}
\includegraphics[width=90mm]{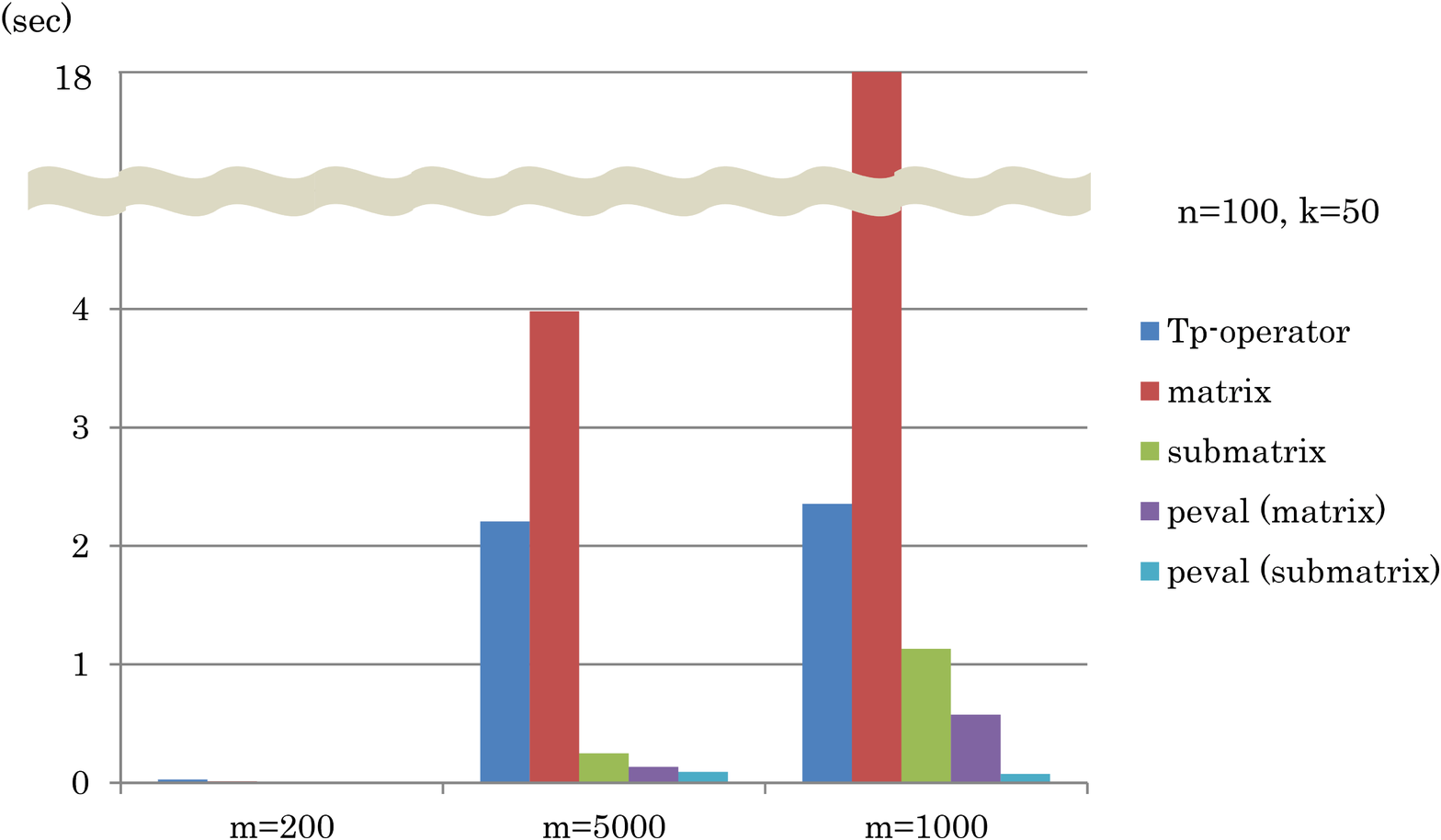} \\
\vspace*{10mm}
\includegraphics[width=90mm]{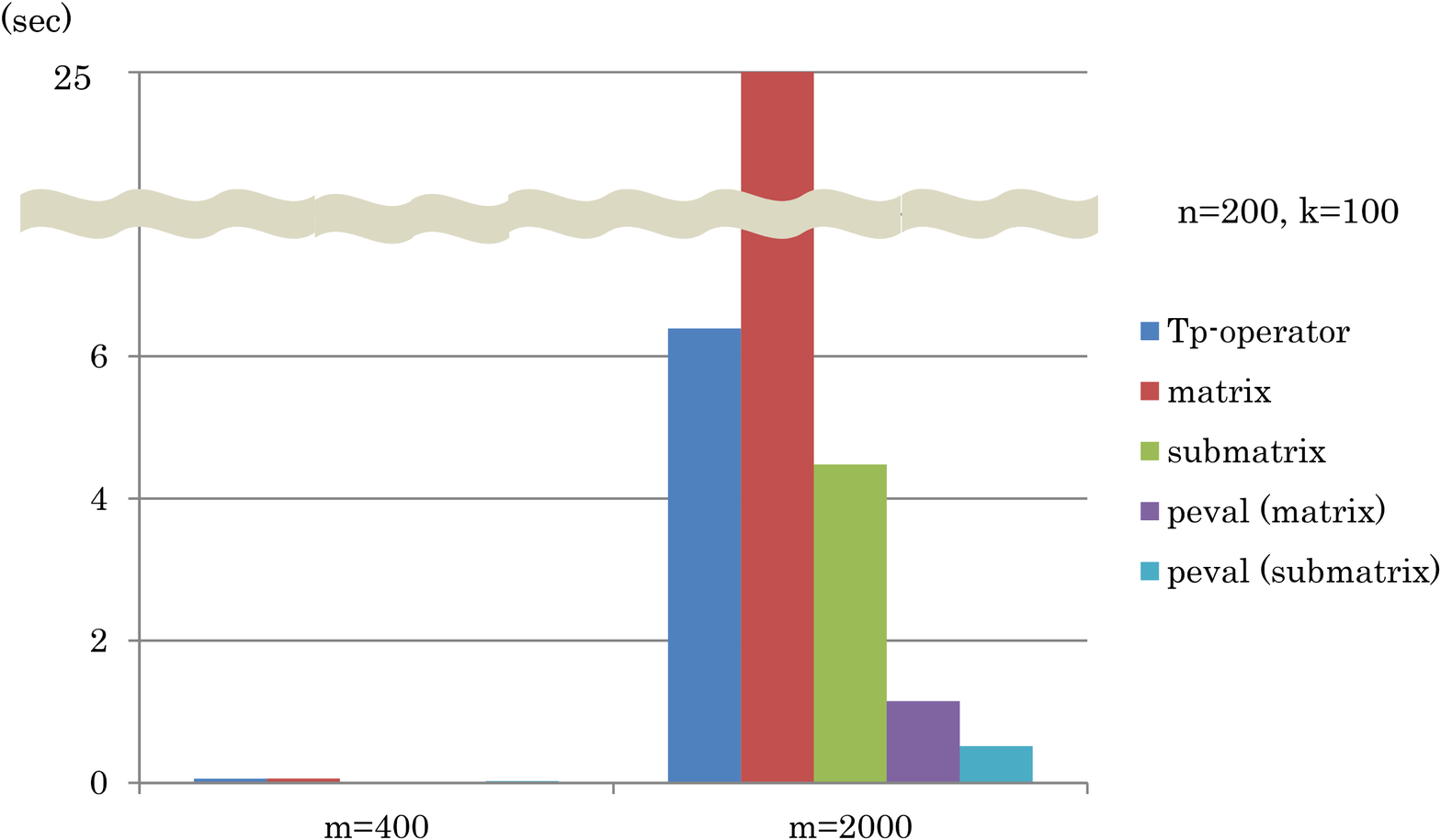} 
\caption{Comparison of runtime for fixpoint computation}\label{fig-ex200}
\end{figure}

We perform experiments by changing parameters $(n,m,k)$.  
For each $(n,m,k)$ we measure runtime at least three times and pick average values. 
Tables~\ref{table-ex50},~\ref{table-ex100} and~\ref{table-ex200} show the results of testing 
for $n=50, 100$ and $200$, respectively. 
In the table,``all" means time for creating a program matrix and computing a fixpoint, and  
``fixpoint" means time for computing a fixpoint. 
In the column of partial evaluation, $\Gamma^k$ means time for partial evaluation, 
``matrix" means fixpoint computation by program matrices after partial evaluation, and 
``col.\ reduct." means fixpoint computation by column reduction after partial evaluation. 
Figure~\ref{fig-ex200} compares runtime for computing fixpoints. 

By those tables, we can observe the following facts. 
\begin{itemize}
\item  For fixpoint computation, 
column reduction outperforms matrix computation and the $T_P$-operator in almost 
every case. Naive computation by program matrices becomes inefficient in large scale of programs. 
\item Column reduction is effective in a large scale of programs. 
It is often more than 10 times faster than naive computation by program matrices. 
\item By performing partial evaluation, time for fixpoint computation is significantly reduced. 
In particular, partial evaluation + column reduction is effective when $k>1$, and 
fixpoint computation by this hybrid method is often more than 10 times faster than other methods 
in large scale of programs (Tables~\ref{table-ex100} and~\ref{table-ex200}). 
\end{itemize}

\begin{table}[h]
\caption{Experimental Results ($n=50$; sec)}\label{table-ex50}
\begin{tabular}{cccccccccc}\hline\hline
\Lower{m} & \Lower{$T_P$-operator} & \multicolumn{2}{c}{matrix} & \multicolumn{2}{c}{column reduction} & \multicolumn{4}{c}{partial evaluation} \\ [1ex] \cline{3-4}\cline{5-6}\cline{7-10} 
&  & fixpoint & all & fixpoint & all & k & $\Gamma^k$ & matrix & col.\ reduct.  \rule{0in}{3ex} \\ \hline 
&  &  &  &  & &  1 & 0.002 & 0.005 & 0.005   \\
100 &  0.008  & 0.007 & 0.155 & 0.005 & 0.13 &   5 & 0.006 & 0.006 & 0.005  \\  
&  &  &  &  & &  25 & 0.006 & 0.005 & 0.005   \\
&  &  &  &  & &  50 & 0.008 & 0.009 & 0.004   \\ \hline 

&  &  &  &  & &  1 & 0.29 & 0.111 & 0.173   \\
1250 &  0.35  & 1.135 & 1.158 & 0.061 & 0.247 &  5 & 0.656 & 0.04 & 0.019  \\  
&  &  &  &  & &  25 & 0.565 & 0.029 & 0.012   \\
&  &  &  &  & &  50 & 0.56 & 0.032 & 0.012   \\ \hline 

&  &  &  &  &                                             &  1 & 0.438 & 0.133 & 0.227   \\
2500 &  0.627  & 1.269 & 1.3 & 0.071 & 0.142 &  5 & 1.41 & 0.066 & 0.05  \\  
&  &  &  &  &                                             &  25 & 1.81 & 0.063 & 0.043   \\
&  &  &  &  &                                             &  50 & 1.401 & 0.067 & 0.06   \\ \hline 

&  &  &  &  & &  1 & 38.938 & 1.142 & 3.189   \\
12500 &  2.081  & 13.937 & 14.358 & 0.649 & 1.024 &  5 & 78.646 & 0.585 & 0.168  \\  
&  &  &  &  & &  25 & 79.625 & 0.604 & 0.168   \\
&  &  &  &  & &  50 & 80.181 & 0.606 & 0.168   \\ \hline 
\end{tabular}
\end{table}
\begin{table}[h]
\caption{Experimental Results ($n=100$; sec)}\label{table-ex100}
\begin{tabular}{cccccccccc}\hline\hline
\Lower{m} & \Lower{$T_P$-operator} & \multicolumn{2}{c}{matrix} & \multicolumn{2}{c}{column reduction} & \multicolumn{4}{c}{partial evaluation} \\ [1ex] \cline{3-4}\cline{5-6}\cline{7-10} 
&  & fixpoint & all & fixpoint & all & k & $\Gamma^k$ & matrix & col.\ reduct.  \rule{0in}{3ex} \\ \hline 
&  &  &  &  &                                              &  1 & 0.007 & 0.006 & 0.009   \\
200 &  0.029  & 0.014 & 0.05 & 0.003 & 0.021 &   5 & 0.018 & 0.007 & 0.007  \\  
&  &  &  &  &                                             &  50 & 0.017 & 0.007 & 0.01   \\
&  &  &  &  &                                              &  100 & 0.026 & 0.008 & 0.009   \\ \hline 

&  &  &  &  &                                                  &  1 & 2.357 & 0.288 & 0.608   \\
5000 &  2.206  & 3.981 & 4.044 & 0.249 & 0.485 &  5 & 5.921 & 0.143 & 0.117  \\  
&  &  &  &  &                                                  &  50 & 6.696 & 0.136 & 0.094   \\
&  &  &  &  &                                                  &  100 & 6.403 & 0.143 & 0.094   \\ \hline 

&  &  &  &  &                                                     &  1 & 38.549 & 1.502 & 1.778   \\
10000 &  2.355  & 18.553 & 18.836 & 1.131 & 1.807   &  5 & 79.68 & 0.603 & 0.075  \\  
&  &  &  &  &                                                     &  50 & 78.037 & 0.576 & 0.076   \\
&  &  &  &  &                                                    &  100 & 77.58 & 0.575 & 0.074   \\ \hline 
\end{tabular}
\end{table}

\begin{table}[h]
\caption{Experimental Results ($n=200$; sec)}\label{table-ex200}
\begin{tabular}{cccccccccc}\hline\hline
\Lower{m} & \Lower{$T_P$-operator} & \multicolumn{2}{c}{matrix} & \multicolumn{2}{c}{column reduction} & \multicolumn{4}{c}{partial evaluation} \\ [1ex] \cline{3-4}\cline{5-6}\cline{7-10} 
&  & fixpoint & all & fixpoint & all & k & $\Gamma^k$ & matrix & col.\ reduct.  \rule{0in}{3ex} \\ \hline 
&  &  &  &  &                                             &  1 & 0.047 & 0.023 & 0.022   \\
400 &  0.06  & 0.063 & 0.075 & 0.013 & 0.06   &   5 & 0.071 & 0.018 & 0.019  \\  
&  &  &  &  &                                             &  100 & 0.102 & 0.018 & 0.024   \\
&  &  &  &  &                                             &  200 & 0.087 & 0.016 & 0.019   \\ \hline 

&  &  &  &  &                                                    &  1 & 138.651 & 2.317 & 6.423   \\
20000 &  6.391  & 25.161 & 25.833 & 4.48 & 7.771 &  5 & 295.462 & 1.173 & 0.529  \\  
&  &  &  &  &                                                    &  100 & 289.564 & 1.15 & 0.519   \\
&  &  &  &  &                                                    &  200 & 285.345 & 1.203 & 0.519   \\ \hline 
\end{tabular}
\end{table}

\section{Concluding Remarks}

In this paper, we introduced a method of embedding logic programs in vector spaces.  
We developed algorithms for computing least models of definite programs, and presented 
column reduction and partial evaluation for optimization. 
Experimental results show that column reduction is effective to realize efficient 
computation in a large scale of programs and 
partial evaluation helps to reduce runtime significantly. 
%
It is known that the least model of a definite program is computed in $O(N)$ \cite{DG84} 
where $N$ is the size (number of literals) of a program.  Since column reduction takes $O(m^2\times n)$ time, it would be effective when $m^2\times n < N$, 
i.e., the size of a program is large with a relatively small number of atoms. 
Moreover, since partial evaluation is performed apart from fixpoint computation, 
combination of column reduction and partial evaluation would be effective in practice.  
The linear algebraic approach enables us to use efficient algorithms of numerical linear algebra and opens perspective for parallel computation of logic programming. 
Performance of our implementation heavily depends on the environment of linear algebraic computation.  
For instance, we could use the CUDA package to accelerate linear algebraic computation on GPU. 
Once more powerful platforms are developed for linear algebraic computation, 
the proposed method would have the merit of such advanced technologies. 
We have used Maple for implementation, but the proposed algorithms can be realized by other programming languages.  We compared runtime in experiments, while it would be interesting to 
compare other metrics in algorithms and matrices that are part of the computation, 
for instance, the number of iterations to the fixpoint, the compression $(m-n)/m$ achieved by column reduction, the sparseness of the matrices with and without partial evaluation, and so on. 

This paper studies algorithms for computing least models of definite programs. 
An important question is whether linear algebraic computation is applied to {\em answer set programming}.  
A method for computing stable models of normal logic programs 
was reported in \cite{SIS17} in which normal programs are represented by third-order tensors. 
Computing large scale of programs in third-order tensors requires scalable techniques and 
optimization, however. 
As an alternative approach, we can use a technique of transforming normal programs 
to definite programs, and computing stable models as least models of the transformed programs  
\cite{AL00}.  Experimental results based on this method are reported in \cite{HSSI18}, and 
partial evaluation would help to reduce runtime.  
We also plan to develop a new algorithm for ASP in vector spaces and evaluate it using benchmark testing. 
Recently, Sato $et\,al.$ \citeyear{Sato18} introduce a method of linear algebraic computation of abduction in Datalog.  
We consider that abductive logic programming would be realized 
in vector spaces by extending the framework introduced in this paper. 
A preliminary result along this line is reported in \cite{ABR18}. 
There is a number of interesting topics to be investigated and rooms for improvement in this new approach to logic programming. 


\end{document}